\documentclass[conference]{IEEEtran}
\usepackage{graphicx}
\usepackage{amsmath,amssymb}
\usepackage{booktabs}
\usepackage{multirow}
\usepackage{tikz}
\usetikzlibrary{positioning, arrows.meta, shapes.geometric}
\usepackage{xcolor}
\usepackage{cite}
\usepackage{algorithm}
\usepackage{algpseudocode}
\usepackage{url}
\usepackage{pgfplots}
\pgfplotsset{compat=1.18} % Or the latest version you have
\usepackage{hyperref}
\usepackage{tabularx}
\usepackage{fontawesome5} % or \usepackage{fontawesome}
\usepackage{soul}
\usepackage{comment}
\usepackage[font=footnotesize, labelfont=bf, labelsep=period]{caption}
\usetikzlibrary{shapes, arrows, positioning, calc, fit, backgrounds}

\title{MARCO: Hardware-Aware Neural Architecture Search for Edge Devices with Multi-Agent Reinforcement Learning and Conformal Prediction Filtering }
\author{%
\IEEEauthorblockN{Arya Fayyazi\IEEEauthorrefmark{1}, Mehdi Kamal\IEEEauthorrefmark{1}, Massoud Pedram\IEEEauthorrefmark{1}}
\IEEEauthorblockA{\IEEEauthorrefmark{1}University of Southern California, Los Angeles, California, US \\
Email: \{afayyazi, mehdi.kamal, pedram\}@usc.edu}}
\begin{document}

\maketitle

\begin{abstract}
This paper introduces MARCO (Multi-Agent Reinforcement learning with Conformal Optimization), a novel hardware-aware framework for efficient neural architecture search (NAS) targeting resource-constrained edge devices. By significantly reducing search time and maintaining accuracy under strict hardware constraints, MARCO bridges the gap between automated DNN design and CAD for edge AI deployment. MARCO's core technical contribution lies in its unique combination of multi-agent reinforcement learning (MARL) with Conformal Prediction (CP) to accelerate the hardware/software co-design process for deploying deep neural networks. Unlike conventional once-for-all (OFA) supernet approaches that require extensive pretraining, MARCO decomposes the NAS task into a hardware configuration agent (HCA) and a Quantization Agent (QA). The HCA optimizes high-level design parameters, while the QA determines per-layer bit-widths under strict memory and latency budgets using a shared reward signal within a centralized-critic, decentralized-execution (CTDE) paradigm. A key innovation is the integration of a calibrated CP surrogate model that provides statistical guarantees (with a user-defined miscoverage rate) to prune unpromising candidate architectures before incurring the high costs of partial training or hardware simulation. This early filtering drastically reduces the search space while ensuring that high-quality designs are retained with a high probability. Extensive experiments on MNIST, CIFAR-10, and CIFAR-100 demonstrate that MARCO achieves a 3–4× reduction in total search time compared to an OFA baseline while maintaining near-baseline accuracy (within 0.3\%). Furthermore, MARCO also reduces inference latency. Validation on a MAX78000 evaluation board confirms that simulator trends hold in practice, with simulator estimates deviating from measured values by less than 5\%.

\end{abstract}

\section{Introduction}
\label{sec:intro}

Edge computing has shifted a substantial fraction of inference workloads from data centers to resource‑constrained devices (microcontrollers, low‑power DSPs, and emerging TinyML accelerators) where on‑chip memory rarely exceeds a few hundred kilobytes and energy budgets restrict inference latency to tens of milliseconds. Early solutions relied on expert‑driven pruning and architecture slimming, leading to streamlined networks such as SqueezeNet~\cite{iandola2016squeezenet}, MobileNet~\cite{howard2017mobilenets}, and ShuffleNet~\cite{zhang2018shufflenet}. As edge applications diversified, designers turned to \emph{hardware‑aware neural‑architecture search} (HWNAS) to automate this process, allowing algorithms, rather than humans, to navigate the vast combinatorial design space while respecting device constraints. Representative milestones include MnasNet's single agent reinforcement learning approach that embeds latency penalties in the reward function~\cite{tan2019mnasnet}, FBNet’s differentiable relaxation that predicts latency through a proxy model~\cite{wu2019fbnet}, and the Once‑for‑All (OFA) supernet paradigm that trains a universe of subnets in a single monolithic graph~\cite{cai2020once}. These paradigms delivered compelling efficiency gains on mobile CPUs and GPUs, but translating them to ultraconstrained edge hardware revealed new bottlenecks.

Despite advances in HWNAS and their potential to reduce inference latency, several limitations hinder their effective deployment on edge devices. OFA‑style supernets demand days of GPU time for pre‑training~\cite{cai2020once}; any change in the target memory or latency budget requires re-specialization, which is incompatible with the rapid iteration cycles typical of computer‑aided design (CAD) flows~\cite{negrinho2017deeparchitect}. Furthermore, most previous searches entangle macro-architecture choices (e.g., layer depth, kernel size, channel width) with quantization policies within a single action space~\cite{cai2020once, wu2019fbnet}; the resultant explosion of dimensions hampers exploration of mixed precision designs that are crucial for squeezing models into sub-megabyte memory footprints~\cite{dong2020hawq}. Moreover, evaluation remains the dominant cost: each sampled sub‑network generally undergoes partial training to estimate accuracy and is then profiled on a hardware simulator~\cite{tan2019mnasnet, lin2020mcunet}, yet a significant fraction ultimately violates resource limits, wasting compute on hopeless candidates. Accelerator‑specific heuristics~\cite{wu2019fbnet, lu2022nasbot} and proxy cost models~\cite{wan2020fbnetv2} mitigate this burden, but do not offer statistical guarantees that near‑optimal architectures survive the pruning stage~\cite{angelopoulos2021gentle}.

To address these challenges, we introduce \textbf{MARCO} (\emph{Multi‑Agent Reinforcement learning with Conformal Optimization}), a hardware-aware NAS framework tailored to a multiplicity of devices with stringent memory and latency constraints. MARCO decomposes the design task into multiple cooperative agents operating under a centralized-critic, decentralized-execution paradigm: for example, a \emph{hardware‑configuration agent} sets macro-level topology parameters, while a \emph{quantization agent} assigns per-layer bitwidths, thereby enabling fine-grained mixed-precision implementation without incurring the combinatorial blow‑up of a flat search. Candidate networks produced by these agents are screened by a calibrated conformal prediction surrogate function that provides distribution‑free confidence upper bounds on reward; architectures whose bounds fall below a user‑defined threshold are rejected \emph{before} any training or simulation. This principled pruning strategy removes roughly one quarter of the search space while guaranteeing, with probability at least \(1-\delta\), that high-quality designs are retained. Extensive experiments on vision benchmarks such as MNIST~\cite{lecun1998gradient}, CIFAR10, and CIFAR100~\cite{krizhevsky2009learning} demonstrate that MARCO cuts the total search time by three to four times relative to the OFA baselines, while maintaining accuracy within 0.3\% of the best previous methods. Because MARCO interfaces only with abstract latency and memory queries supplied by the target toolchain, it remains portable across a spectrum of microcontrollers and (low-end) edge accelerators.

This work claims several key novelties:
\begin{enumerate}
\item It presents the first multi-agent hardware-aware NAS (HW-NAS) formulation that explicitly decouples architecture optimization and data quantization for edge devices.
\item It introduces a conformal prediction filter that offers provable coverage guarantees while drastically reducing the evaluation cost.
\item It delivers a modular hardware implementation that integrates seamlessly with standard TinyML toolchains by interfacing only with abstract latency and memory queries to achieve state‑of‑the‑art search efficiency on various FPGA devices.
\end{enumerate}

%The remainder of the paper is organized as follows. Section~\ref{sec:preliminaries} reviews foundational concepts in multi‑agent reinforcement learning, conformal prediction, and prior HWNAS literature. Section~\ref{sec:method} details the MARCO methodology, including agent policies, reward shaping, and conformal calibration. The experimental results and ablation studies are reported in Section~\ref{sec:experiments}, and finally, the article is concluded in Section~\ref{sec:conclusion}.

\section{Preliminaries and Related Work}
\label{sec:preliminaries}

This section provides a comprehensive background on the key concepts and literature that form the foundation of our proposed \textbf{MARCO} (\underline{MAR}l + \underline{C}onformal \underline{O}ptimization) approach for HWNAS on microcontrollers. We first introduce the fundamentals of multi-agent reinforcement learning (MARL) and the centralized training–decentralized execution (CTDE) paradigm, which effectively decouples high-dimensional design decisions. We then present the principles of Conformal Prediction (CP), emphasizing its capability to provide rigorous coverage guarantees for resource-constrained applications. Finally, we review and compare related NAS and CAD-oriented techniques, including MCUNet, MnasNet, FBNet, ProxylessNAS, and AutoTVM, as well as complementary toolchains such as MaximAI and HLS4ML, highlighting how MARCO differs by integrating multi-agent RL with CP-based early filtering under strict hardware constraints.

\subsection{Multi-Agent Reinforcement Learning (MARL) and CTDE}
Standard reinforcement learning (RL) typically involves a single agent that learns to maximize a cumulative reward by interacting with an environment \cite{sutton2018reinforcement}. However, complex design tasks (such as HWNAS) often entail managing both macro-level architectural decisions (e.g., layer counts, kernel sizes, channel widths) and micro-level parameters (e.g., bit-width quantization). In this context, a single agent must operate over an intractably large action space. Multi-agent reinforcement learning (MARL) alleviates this problem by decomposing the decision space among multiple agents, each responsible for a subset of the overall action space \cite{gronauer2022multi, yang2020overview}. 

A widely adopted MARL framework is the paradigm \emph{centralized training–decentralized execution} (CTDE) \cite{foerster2018counterfactual, yu2022surprising, carmack2021review}. During training, a centralized critic observes the complete state of the environment, facilitating stable policy updates for each agent (or actor) that only has access to local observations during execution. In the context of HWNAS, a Hardware Configuration Agent (HCA) selects macro-level design parameters, while a Quantization Agent (QA) determines the per-layer bit-width settings. Through CTDE, the critic assesses the entire partial architecture, taking into account cumulative memory usage, partial accuracy from fast training, and hardware constraints (e.g., a memory limit \(M_{\max}\)), to provide a cohesive and hardware-oriented reward signal. This enables each agent to adapt its policy using local observations while ensuring global feasibility.

\subsection{Conformal Prediction (CP) for Resource-Constrained Systems}
Conformal Prediction (CP) is a distribution-free statistical framework that produces predictive intervals with a guaranteed coverage probability of at least \(1-\delta\) \cite{shafer2008tutorial, angelopoulos2021gentle,fayyazi2025facter}. In a regression setting, a surrogate model \(g(\cdot)\) is trained to predict a scalar target (here, the reward \(R(a)\)) from a feature vector \(x(a)\) that describes an architecture. Given a calibration set \(\{(a_i, R(a_i))\}_{i=1}^{M}\), CP computes the residuals 
\begin{equation}
\varepsilon_i = \left| R(a_i) - g(x(a_i)) \right|,
\end{equation}
and defines an uncertainty offset \(\alpha_{1-\delta}\) as the \((1-\delta)\)-quantile of these residuals. With this setup, CP ensures that with probability at least \(1-\delta\),
\begin{equation}
R(a) \in \left[ g(x(a)) - \alpha_{1-\delta},\, g(x(a)) + \alpha_{1-\delta} \right].
\end{equation}
In our implementation, this property is used to preemptively discard candidate architectures that are unlikely to achieve a reward above a threshold \(\tau\). If the upper bound \(g(x(a))+\alpha_{1-\delta}\) falls below \(\tau\), the candidate is not further evaluated, saving costly partial training and simulation time, a crucial advantage in resource-constrained CAD environments.

\subsection{TinyML NAS: Approaches and CAD-Oriented Toolchains}
The recent surge in TinyML has spurred the development of NAS frameworks tailored for microcontrollers and other highly resource-constrained devices. Conventional approaches such as the Once-for-All (OFA) supernet strategy~\cite{cai2020once}, as implemented in the MaximAI toolchain, require extensive supernet training over many days before viable sub-networks can be extracted.
Although effective, this method is computationally expensive and impractical for rapid design iterations.

Alternative approaches include:
\begin{itemize}
    \item \textbf{MCUNet (TinyNAS)}~\cite{lin2020mcunet}, which uses evolutionary search and pruning to optimize architectures for MCUs, but relies on fixed 8-bit quantization and does not leverage multi-agent decoupling.
    \item \textbf{MnasNet}~\cite{tan2019mnasnet}, a single-agent RL technique originally designed for mobile devices, focuses on 8-bit uniform quantization and is not explicitly optimized for microcontrollers with stringent memory limits.
    \item \textbf{FBNet}~\cite{wu2019fbnet} and \textbf{ProxylessNAS}~\cite{cai2019proxylessnas} represent gradient-based NAS approaches that incorporate hardware cost models, yet typically target platforms with higher memory budgets and do not exploit multi-agent methods to decouple architecture and quantization decisions.
    \item \textbf{AutoTVM}~\cite{chen2018tvm} and related works~\cite{fayyazi2025dcoc} address kernel-level optimization and operator scheduling on high-performance devices rather than the entire DNN design space, focusing less on mixed-precision or memory fragmentation issues.
\end{itemize}

Moreover, CAD-oriented toolchains such as \textbf{MaximAI} and \textbf{HLS4ML}~\cite{duarte2018hls4ml} have advanced the automation of hardware/software co-design. Still, they generally rely on monolithic supernet training or fixed-design spaces, lacking the flexibility to adjust bit-widths and other design knobs through reinforcement learning dynamically. In contrast, MARCO is novel in its integration of multi-agent RL with CP-based early filtering; this combination not only reduces search time by discarding low-potential architectures but also enforces strict hardware constraints such as memory and latency, making it highly suitable for CAD-driven microcontroller design.

Furthermore, the foundational work in hardware-aware design and FPGA-based acceleration by Cong et al.~\cite{zhang2015optimizing, wang2021soda} has demonstrated the importance of integrating design space exploration with hardware performance metrics. Our approach builds on these CAD principles by embedding explicit memory and latency constraints within the reward function and leveraging statistical filtering (via CP) to minimize computational overhead.

In summary, while existing methods such as MCUNet, MnasNet, FBNet, and ProxylessNAS have made significant strides in efficient network design, they typically do not combine multi-agent policy gradients with CP filtering. MARCO achieves significantly reduced search times and improved resource utilization and bridges the gap between automated DNN design and CAD for edge AI deployment.

\section{Methodology}
\label{sec:method}

%%%%%%%%%%%%%%%%%%%%%%%%%%%%%%%%%%%%%%%%%%%%%%%%%%%%%%%%%%%%%%%%%%%%%%%%
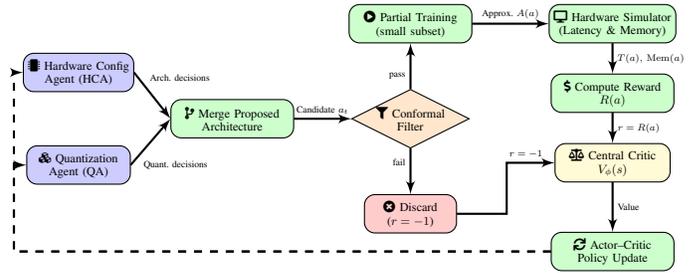
\begin{figure}[t]
\centering
\begin{tikzpicture}[scale=0.51, transform shape,
    node distance=1.4cm and 1.8cm, 
    auto,
    >=latex',
    font=\small,
    arrow/.style={->, thick},
    darrow/.style={->, thick, dashed},
    agent/.style={
       rectangle, 
       draw, 
       rounded corners, 
       fill=blue!20,
       minimum width=2.7cm, 
       minimum height=1.0cm,
       align=center
    },
    block/.style={
       rectangle, 
       draw, 
       rounded corners, 
       fill=green!20,
       minimum width=3.2cm, 
       minimum height=1.0cm,
       align=center
    },
    diamondbox/.style={
       diamond,
       aspect=2,
       draw,
       fill=orange!20,
       align=center,
       minimum width=2.6cm,
       minimum height=1.2cm,
       inner sep=0pt
    },
    disc/.style={
       rectangle,
       draw,
       rounded corners,
       fill=red!20,
       minimum width=2.4cm, 
       minimum height=1.0cm,
       align=center
    },
    critic/.style={
       rectangle,
       draw,
       rounded corners,
       fill=yellow!20,
       minimum width=3.0cm,
       minimum height=1.0cm,
       align=center
    }
]

% Agents
\node[agent] (HCA) {\faMicrochip\ Hardware Config\\Agent (HCA)};
\node[agent, below=1.4cm of HCA] (QA) {\faCubes\ Quantization\\Agent (QA)};

% Merge
\node[block, right=2.4cm of $(HCA)!0.5!(QA)$] (Merge) {\faCodeBranch\ Merge Proposed \\ Architecture};

% CP Filter
\node[diamondbox, right=1.5cm of Merge] (CP) {\faFilter\ Conformal\\Filter};

% Discard
\node[disc, below=1.2cm of CP] (Discard) {\faTimesCircle\ Discard\\($r=-1$)};

% Partial Training
\node[block, above=1.2cm of CP] (PartialTrain) {\faPlayCircle\ Partial Training\\(small subset)};

% Simulator
\node[block, right=2.0cm of PartialTrain] (Sim) {\faDesktop\ Hardware Simulator\\(Latency \& Memory)};

% Reward Computation
\node[block, below=0.8cm of Sim] (Reward) {\faDollarSign\ Compute Reward\\$R(a)$};

% Critic
\node[critic, below=0.8cm of Reward] (Critic) {\faBalanceScale\ Central Critic\\$V_{\phi}(s)$};

% Policy update
\node[block, below=1.3cm of Critic] (Update) {\faSync\ Actor–Critic\\Policy Update};

% ARROWS
\draw[arrow] (HCA.east) -- node[above right,pos=0.3]{\scriptsize Arch.\ decisions} (Merge.west);
\draw[arrow] (QA.east) -- node[below right,pos=0.2]{\scriptsize Quant.\ decisions} (Merge.west);
\draw[arrow] (Merge.east) -- node[above]{\scriptsize Candidate $a_t$} (CP.west);
\draw[arrow] (CP.north) -- node[left,pos=0.3]{\scriptsize pass} (PartialTrain.south);
\draw[arrow] (CP.south) -- node[left,pos=0.3]{\scriptsize fail} (Discard.north);
\draw[arrow] (PartialTrain.east) -- node[above]{\scriptsize Approx.\ $A(a)$} (Sim.west);
\draw[arrow] (Sim.south) -- node[right]{\scriptsize $T(a),\,\mathrm{Mem}(a)$} (Reward.north);
\draw[arrow] (Discard.east) -- ++(1.3,0) |- node[above,pos=0.7]{\scriptsize $r=-1$} (Critic.west);
\draw[arrow] (Reward.south) -- node[right]{\scriptsize $r=R(a)$} (Critic.north);
\draw[arrow] (Critic.south) -- node[right]{\scriptsize Value} (Update.north);
\draw[darrow] (Update.west) -- ++(-14,0) |- (HCA.west);
\draw[darrow] (Update.west) -- ++(-14,0) |- (QA.west);

\end{tikzpicture}
\caption{MARCO workflow. At each time step, the agents (\faMicrochip\ HCA and \faCubes\ QA) propose a candidate, which is merged and sent to a \faFilter\ CP filter. If the upper confidence bound is below threshold $\tau$, it is \faTimesCircle\ discarded with $r=-1$. Otherwise, \faPlayCircle\ partial training estimates accuracy, \faDesktop\ a hardware simulator returns latency and memory, then \faDollarSign\ reward computation and \faBalanceScale\ centralized critic updates converge policies via \faSync\ actor–critic steps.}
\label{fig:flow_advanced}
\end{figure}

This section details the proposed \textbf{MARCO} (MARl + Conformal Optimization) framework for HWNAS on memory-constrained edge platforms. MARCO formulates the NAS problem as a series of coordinated decisions made by multiple reinforcement learning (RL) agents within a centralized training–decentralized execution (CTDE) paradigm. It incorporates Conformal Prediction (CP) filtering to discard suboptimal candidate architectures before incurring expensive evaluation costs. 

Our workflow (see Figure~\ref{fig:flow_advanced}) consists of the following steps. In each RL episode, two specialized agents, the Hardware Configuration Agent (HCA) and the Quantization Agent (QA), independently decide on macro-architecture parameters (layer count, kernel sizes, channel widths, skip/pooling choices) and per-layer quantization (e.g., 4-bit or 8-bit), respectively. Their actions are merged to form a candidate architecture $a_t$. A Conformal Prediction filter then evaluates the candidate’s predicted reward using a surrogate model and, if the candidate’s upper confidence bound falls below a predetermined threshold, the candidate is discarded. Otherwise, the candidate undergoes partial training and simulation to obtain an estimated accuracy $A(a_t)$, measured latency $T(a_t)$, and memory usage $\mathrm{Mem}(a_t)$, which are combined into a reward. Finally, a centralized critic computes value estimates from the partially defined state and assists in updating the policies of both agents via Proximal Policy Optimization (PPO).

\subsection{Design Knobs and Search Space}
\label{subsec:design_knobs}

The design space $\mathcal{A}$ is parametrized by several hardware-aware design knobs that directly affect a candidate network's memory footprint, latency, and predictive accuracy. Table~\ref{tab:design_knobs} summarizes these knobs along with their fixed choices and rationales. For example, using a 4-bit quantization for selected layers reduces memory usage by exactly 50\% compared to 8-bit, while increasing channel widths or the number of layers improves accuracy at the expense of additional memory and latency. In our framework, these design knobs are deterministically set and embedded in the action spaces of the agents.

Although our experimental demonstrations adopt configurations inspired by a specific low-power microcontroller equipped with a DNN accelerator, the MARCO framework is fully general. It can be applied to any platform with well-defined latency and memory constraints. The design knobs and simulation setup used here are easily adjustable to suit a wide range of resource-limited hardware architectures in both embedded and on-device AI domains.

\begin{table}[t]
\centering
\caption{Key design knobs in \textbf{MARCO} and their rationale (derived from a representative low-power DNN accelerator). These knobs are easily adjustable for other hardware.}
\label{tab:design_knobs}
\small
\resizebox{\linewidth}{!}{%
\begin{tabular}{l p{2.6cm} p{4.4cm}}
\toprule
\textbf{Knob} & \textbf{Fixed Choices} & \textbf{Impact and Rationale} \\
\midrule
Number of Layers & \small MobileNet-like: 4--12; ResNet-like: 8--20 & Deeper networks can improve accuracy but incur higher latency and memory usage. \\
Kernel Size      & \small $3\times3$ and $5\times5$ & A smaller kernel is efficient; a larger kernel provides more receptive field at higher cost. \\
Channel Width    & \small 8, 16, 32, 64, 128 & Wider channels boost capacity at a quadratic increase in mem. usage. \\
\begin{tabular}{@{}c@{}}Skip Connections / \\ Pooling\end{tabular}
  & \texttt{"skip"} or \texttt{"no skip"}; \texttt{"max pooling"} or \texttt{"average pooling"} & Influences feature propagation and intermediate buffer sizes. \\
Per-layer Bit-width & \small 4-bit and 8-bit & Lower bit-width reduces memory and latency, critical under small  $\mathrm{Mem}_{budget}$. \\
\bottomrule
\end{tabular}%
}
\end{table}

\subsection{CTDE Multi-Agent Reinforcement Learning}
\label{subsec:ctde_marl}

Our method employs a CTDE-based multi-agent RL architecture. Since in this work, MARCO has two agents, we denote the actions of the two agents as:
\begin{equation}
a_t = \bigl(a_t^1,\, a_t^2\bigr),
\end{equation}
where the Hardware Configuration Agent (HCA) selects macro-level design parameters and the Quantization Agent (QA) chooses the bit-widths for individual layers.

\paragraph{State Space}  
At each discrete time step $t$, the global state is defined as:
\begin{equation}
\begin{aligned}
s_t = \Bigl\{\, 
    &\bigl\{ (\text{kernel size}_i,\, \text{channel width}_i,\, \text{bit-width}_i)\bigr\}_{i=1}^{n}, \\
    &\mathrm{Mem}_{\text{used}},\quad \mathrm{Acc}_{\text{partial}},\quad 
    \mathrm{Mem}_{\text{budget}},\quad T_{\mathrm{budget}} 
\Bigr\}
\end{aligned}
\end{equation}
where $n$ is the number of layers defined so far, $\mathrm{Mem}_{\text{used}}$ is the cumulative memory consumption (factoring the effect of 4-bit versus 8-bit quantization), and $\mathrm{Acc}_{\text{partial}}$ is the accuracy estimate obtained from partial training. 
The memory budget is defined by $\mathrm{Mem}_{budget}$ (e.g., 512Kb), and the latency budget is defined by $T_{\mathrm{budget}}$ (e.g., $10\,$ms for CIFAR-10).

\paragraph{Action Spaces}  
The HCA ($\pi_{\theta_1}$) produces a macro‑level decision
\begin{equation}
a_t^1 \sim \pi_{\theta_1}(a_t^1 \mid o_t^1),
\end{equation}
determining design parameters such as layer count, kernel size, channel width, and whether skip connections or pooling is necessary.  Simultaneously, the QA ($\pi_{\theta_2}$) selects
\begin{equation}
a_t^2 \sim \pi_{\theta_2}(a_t^2 \mid o_t^2),
\end{equation}
which defines the bit‑width for the layer currently under construction.  
The \emph{local observations} $o_t^i$ furnish each agent with just the information it needs:  
for the HCA, $o_t^1$ concatenates the depth built so far, kernel and channel choices of previous layers, cumulative memory usage, and the layer index;  
For the QA, $o_t^2$ includes the same cumulative memory counter, the kernel and channel width of the \emph{current} layer (chosen by the HCA), a one‑hot position flag, and the running estimate of partial accuracy.  
Neither agent sees the other’s private action. Yet, both can infer global feasibility through the shared memory and latency fields, enabling decentralized execution while remaining aligned with the overall resource budget.

\paragraph{Reward Function}  
After merging the actions into a candidate architecture $a_t$, the candidate is evaluated with \textit{partial training}, a fast proxy in which the network is fine‑tuned for 5 epochs on a fixed 10\% calibration subset using a frozen learning‑rate schedule.  This inexpensive procedure provides a stable estimate of top‑1 accuracy \(A(a_t)\), while it costs less than 5\% of full training time.  
The reward is then computed as
\begin{equation}
\label{eq:reward}
R(a_t)=A(a_t)-\lambda\,\frac{T(a_t)}{T_{\mathrm{budget}}}-\mu\,\mathbf{1}\{\mathrm{Mem}(a_t)>M_{\mathrm{budget}}\},
\end{equation}
where \(T(a_t)\) is the latency in milliseconds reported by the simulator, and the indicator penalizes memory overflow.  
The coefficient \(\lambda\) (empirically 0.2) balances latency against accuracy, and an energy term is omitted because power is fixed on constant‑clock embedded hardware.

\paragraph{PPO‑based Actor-Critic Update}  
Both agents are trained with Proximal Policy Optimization (PPO)~\cite{schulman2017ppo}.  
At each step, the environment returns a scalar \emph{reward} \(r_t\) calculated from latency, memory, and accuracy (Eq.~\ref{eq:reward}) and a \emph{global state} \(s_t\) that encodes the partially built architecture, cumulative resource usage, and current latency budget (see Sect.~\ref{subsec:ctde_marl}).  
PPO updates each agent by comparing its \emph{new} action probability with the \emph{old} one:  

\begin{equation}
r_t(\theta_i)=\frac{\pi_{\theta_i}(a_t^i\mid o_t^i)}{\pi_{\theta_i}^{\text{old}}(a_t^i\mid o_t^i)},
\end{equation}

\noindent a likelihood ratio sometimes called the \emph{importance weight}.  
Here \(r_t\) (environment reward) guides value‑function targets, while \(r_t(\theta_i)\) (probability ratio) scales the advantage inside the clipped surrogate loss.  

\begin{equation}
\label{eq:ppo_loss}
\mathcal{L}^{\text{PPO}}(\theta_i)=
\mathbb{E}_t\!\Big[\!
\min\!\big(
r_t(\theta_i)A_t,\,
\mathrm{clip}(r_t(\theta_i),1-\epsilon,1+\epsilon)A_t
\big)
\Big],
\end{equation}

\noindent with advantage \(A_t=r_t+\gamma V_\phi(s_{t+1})-V_\phi(s_t)\), clipping parameter \(\epsilon\), and discount \(\gamma\).  
The centralized critic \(V_\phi\) is updated by least‑squares regression to the one‑step TD target \(\,r_t+\gamma V_\phi(s_{t+1})\).

\subsection{Conformal Prediction (CP) Filter}
\label{subsec:cp_filter}

To accelerate the search process, we integrate a Conformal Prediction (CP) module to preemptively discard candidate architectures unlikely to yield a reward above a predetermined threshold. The CP filter operates by fitting a surrogate regression model $g(x(a))$ to predict the reward $R(a)$ from a feature vector $x(a)$ that characterizes the candidate, while providing rigorous statistical guarantees.

\subsubsection{Theoretical Guarantees}
\label{subsubsec:cp_theory}

Conformal Prediction (CP) provides distribution-free coverage guarantees for the predicted reward intervals. Given a calibration set $\{(a_i, R(a_i))\}_{i=1}^M$ of previously evaluated architectures, we compute residuals $\varepsilon_i = |R(a_i) - g(x(a_i))|$ and define the uncertainty offset $\alpha_{1-\delta}$ as the $(1-\delta)$-quantile of these residuals. By construction, CP guarantees that for any new candidate $a$, the true reward satisfies:
\begin{equation}
    \mathbb{P}\left[R(a) \leq g(x(a)) + \alpha_{1-\delta}\right] \geq 1 - \delta,
    \label{eq:cp_coverage}
\end{equation}
where $\delta \in (0,1)$ is the user-specified miscoverage rate \cite{angelopoulos2021gentle,shafer2008tutorial}. 
In MARCO, we set $\delta = 0.1$, ensuring that with 90\% probability, the upper confidence bound (UCB) $g(x(a)) + \alpha_{1-\delta}$ contains the true reward. The filtering rule:
\begin{equation}
    \text{Discard candidate } a \iff g(x(a)) + \alpha_{1-\delta} < \tau,
    \label{eq:cp_rule}
\end{equation}
implies that at most $\delta$ fraction of viable candidates (those with $R(a) \geq \tau$) will be incorrectly pruned. This trade-off allows for aggressive computational savings while bounding accuracy loss.

\paragraph{Assumptions and Practical Considerations} 
The Eq.~\eqref{eq:cp_coverage} guarantee is valid under the exchangeability assumption (i.e., calibration and test candidates are exchangeable). While MARCO's iterative policy updates could theoretically violate this, we mitigate drift by periodically recalibrating $g(\cdot)$ during training. 
% \textcolor{olive}{In this work, the calibration set size ($M$) was set to $100$ to ensure stable quantile estimation, with empirical convergence of $\alpha_{1-\delta} = 0.8$ \st{observed across trials.}} 
% ==========================================

\subsubsection{Surrogate Model and Feature Vector}
\label{subsubsec:surrogate}

The feature vector $x(a)$ includes:
\begin{equation}
\begin{aligned}
x(a) = \bigl[\, 
  &\text{layer count},\quad \text{channel widths (one-hot encoding)}, \\[1mm]
  &\text{quantization bit-widths},\quad \text{partial accuracy},\ \dots 
\bigr]
\end{aligned}
\end{equation}

We implement $g(\cdot)$ as a \textbf{3-layer MLP} with 64 hidden units per layer and ReLU activations. It is trained on a calibration set $\{(x(a_i), R(a_i))\}_{i=1}^{M}$ using mean squared error loss. Algorithm~\ref{alg:cp_calibrate} furthermore illustrates the pseudocode for the calibration of CP.

\subsubsection{CP Filtering Rule}
\label{subsubsec:filter_rule}

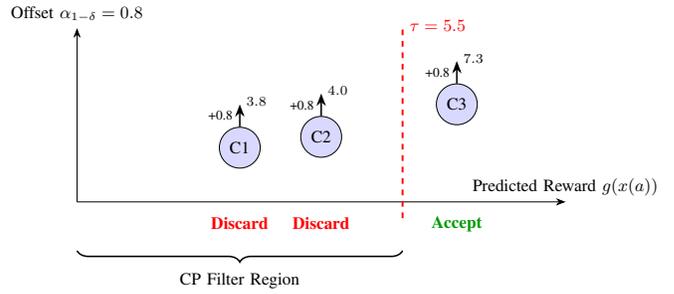
\begin{figure}[t]
\centering
\resizebox{\linewidth}{!}{%
\begin{tikzpicture}[>=Stealth, node distance=1.2cm and 1.2cm, every node/.style={font=\small}]
    % Axes
    \draw[->] (0,0) -- (9,0) node[above] {\small Predicted Reward $g(x(a))$};
    \draw[->] (0,0) -- (0,3.2) node[above] {\small Offset $\alpha_{1-\delta} = 0.8$};

    % Threshold line
    \draw[dashed, line width=1pt, red] (6,-0.3) -- (6,3.2) node[pos=0.95, anchor=south west, font=\small\color{red}] {$\tau = 5.5$};

    % Candidates
    \node[draw, circle, fill=blue!15] (cand1) at (3,1) {C1};
    \node[draw, circle, fill=blue!15] (cand2) at (4.5,1.2) {C2};
    \node[draw, circle, fill=blue!15] (cand3) at (7,1.8) {C3};

    % Offset arrows
    \draw[->, thick] (cand1) -- ++(0,0.8) node[midway,left] {\scriptsize +0.8};
    \draw[->, thick] (cand2) -- ++(0,0.8) node[midway,left] {\scriptsize +0.8};
    \draw[->, thick] (cand3) -- ++(0,0.8) node[midway,left] {\scriptsize +0.8};

    % Result values
    \node at ($(cand1)+(0,0.85)$) [right] {\scriptsize $3.8$};
    \node at ($(cand2)+(0,0.85)$) [right] {\scriptsize $4.0$};
    \node at ($(cand3)+(0,0.85)$) [right] {\scriptsize $7.3$};

    % Decision labels
    \node at (3,-0.4) [text=red, font=\small\bfseries] {Discard};
    \node at (4.5,-0.4) [text=red, font=\small\bfseries] {Discard};
    \node at (7,-0.4) [text=green!60!black, font=\small\bfseries] {Accept};

    % Brace
    \draw [decorate, decoration={brace, amplitude=6pt, mirror}, thick] (0,-0.9) -- (6,-0.9) 
        node[midway, below=8pt]{\small CP Filter Region};
\end{tikzpicture}
} % end resizebox
\caption{CP Filtering mechanism. For each candidate $a$, its predicted reward $g(x(a))$ is increased by the empirically obtained offset $\alpha_{1-\delta}=0.8$. Candidates with $g(x(a))+0.8 < 5.5$ (as seen for C1: $3+0.8=3.8$ and C2: $4+0.8=4.8$) are discarded, while candidates above the threshold (e.g., C3: $6.5+0.8=7.3$) are accepted.}
\label{fig:cp_filter}
\end{figure}

Following CP theory \cite{angelopoulos2021gentle}, with probability at least $1-\delta$, the true reward satisfies:
\begin{equation}
R(a) \le g(x(a)) + \alpha_{1-\delta}.
\end{equation}
Therefore, in this work, if for a new candidate we have
\begin{equation}
g(x(a)) +  \alpha_{1-\delta} < \tau,
\end{equation}
With a threshold $\tau$, the candidate is predicted to be suboptimal and is discarded by assigning a reward $r=-1$, skipping any further evaluation (i.e., partial training or simulator call). Figure~\ref{fig:cp_filter} illustrates the CP filtering mechanism.  

% \subsubsection{Pseudocode for CP Calibration}
% \label{subsec:cp_calibration}
\begin{algorithm}[t]
\small
\caption{CP Calibration: Residual Computation and Quantile Selection}
\label{alg:cp_calibrate}
\begin{algorithmic}[1]
\Require Calibration set $\{(a_i, R(a_i))\}_{i=1}^{M}$, target coverage $1-\delta$
\State Train surrogate model $g(\cdot)$ on $\{(x(a_i), R(a_i))\}$ using MSE loss.
\For{$i=1$ to $M$}
    \State $\varepsilon_i \gets \bigl| R(a_i) - g(x(a_i)) \bigr|$
\EndFor
\State Determine $\alpha_{1-\delta}$ as the $(1-\delta)$ quantile of $\{\varepsilon_1, \varepsilon_2, \ldots, \varepsilon_M\}$
\State \Return $g(\cdot)$ and $\alpha_{1-\delta}$
\end{algorithmic}
\end{algorithm}

\subsection{Algorithmic Integration of MARL and CP}
\label{subsec:alg_flow}

Algorithm~\ref{alg:fullmethod} provides the complete pseudocode for MARCO, integrating multi-agent RL with CP filtering. The surrogate model predicts the reward for each candidate generated by the HCA and QA. If the CP condition $ g(x(a_t)) + \alpha_{1-\delta} < \tau $ holds, the candidate is immediately assigned $r_t = -1$ (discarded) and no further evaluation is performed. Otherwise, the candidate undergoes exactly $E$ epochs of partial training to obtain $A(a_t)$ and is passed to the hardware simulator to measure latency $T(a_t)$ and memory usage $\mathrm{Mem}(a_t)$. Finally, the reward is computed using Eq.~\eqref{eq:reward}, and the centralized critic updates are performed via PPO. 

\begin{algorithm}[t]
\small
\caption{MARCO: Multi-Agent RL with Conformal Filtering}
\label{alg:fullmethod}
\begin{algorithmic}[1]
\Require $\mathrm{Mem}_{budget}$, $T_{\mathrm{budget}}$,$\tau$, $\delta$, $M$, $\lambda$, $\mu$.
\State \textbf{Initialize:}  $\pi_{\theta_1}$, $\pi_{\theta_2}$, $V_{\phi}$.
\State \textbf{Obtain} calibration set $\{(a_i, R(a_i))\}_{i=1}^{100}$ via initial exploration.
\State \textbf{Fit} surrogate $g(\cdot)$ on $\{(x(a_i), R(a_i))\}$ and compute $\alpha_{1-\delta}$ using Algorithm~\ref{alg:cp_calibrate}.
\For{episode = 1 to $N_{\mathrm{ep}}$}
  \For{trial = 1 to $K$}
    \State Observe global state $s_t$.
    \State HCA generates $a_t^1 \sim \pi_{\theta_1}(\cdot \mid o_t^1)$.
    \State QA generates $a_t^2 \sim \pi_{\theta_2}(\cdot \mid o_t^2)$.
    \State Merge to form candidate $a_t=(a_t^1, a_t^2)$.
    \State Compute $\hat{R}_t = g(x(a_t))$.
    \If{$\hat{R}_t + \alpha_{1-\delta} < \tau$}
      \State Set $r_t \gets -1$ \Comment{Discard candidate; no partial training/simulation.}
    \Else
      \State Conduct $E$ epochs of partial training to obtain $A(a_t)$.
      \State Query the HW simulator to find $T(a_t)$ and $\mathrm{Mem}(a_t)$.
      \State Compute reward by using Eg.~\ref{eq:reward}.
      %\[
      %r_t = A(a_t) - \lambda\cdot\frac{T(a_t)}{T_{budget}} - \mu\cdot\mathbf{1}\{\mathrm{Mem}(a_t)> Mem_{budget} \}.
      %\]
    \EndIf
    \State Compute TD error: $\delta_t = r_t + \gamma\,V_{\phi}(s_{t+1}) - V_{\phi}(s_t)$.
    \State Update critic $V_{\phi}$ by minimizing $\delta_t^2$.
    \State Compute advantage $A_t = \delta_t$.
    \State Update each actor $\pi_{\theta_i}$ using the PPO loss (Eq.~\ref{eq:ppo_loss}).
    \State Optionally, re-fit $g(\cdot)$ with the new pair $(x(a_t), R(a_t))$.
  \EndFor
\EndFor
\end{algorithmic}
\end{algorithm}

\section{Results and Discussion}
\label{sec:experiments}

To demonstrate \textbf{MARCO}’s effectiveness and portability we perform two complementary evaluations.  
First, an extensive \emph{simulator‑based study} (using a low‑power microcontroller simulator that exposes cycle‑accurate latency and SRAM usage) establishes search‑time, accuracy, and memory trade‑offs under controlled conditions.  
Second, a concise \emph{real‑hardware validation} deploys the highest‑reward architectures onto the corresponding evaluation board, confirming that simulator trends hold in practice.  
The microcontroller chosen for both steps possesses a dedicated CNN accelerator and a sub‑megabyte SRAM budget; it serves as a representative proxy for contemporary resource‑constrained edge devices. All search knobs (Table~\ref{tab:design_knobs}) and reward budgets derive from this chip’s public datasheet, yet \emph{every quantity is user‑configurable}, allowing MARCO to retarget other edge platforms with different memory or latency limits by a single JSON file change.

\subsection{Simulator‑Based Evaluation}
\label{sec:sim_results}

\subsubsection{Hardware Configuration and Memory Analysis}
\label{sec:exp:hardware}

\paragraph{MAX78000 MCU Setup}
Our experiments are performed on the \textbf{MAX78000} microcontroller, clocked at 100 MHz, featuring 512 kB of on-chip SRAM dedicated to CNN weights and 160 kB of SRAM for activations and intermediate buffers \cite{max78000_datasheet}.  The MAX78000 incorporates a dedicated CNN accelerator that supports 8‑bit fixed‑point inference with optional 4‑bit mixed‑precision operation \cite{max78000_appnote}.  We use the official Analog Devices MaximAI simulator to estimate inference latency \(T(a)\) and verify that candidate architectures satisfy \(\mathrm{Mem}(a)\le512\)\,kB \cite{maximai_toolkit}. The selected networks are then flashed onto the MAX78000EVKIT to confirm that the simulated performance matches on‑device measurements.

\paragraph{SRAM/Flash Usage and Fragmentation}
Final weights are stored in Flash and then mapped to SRAM at runtime. Partial 4-bit quantization can result in irregular layer sizes and potential SRAM fragmentation. Our environment penalizes or discards architectures that cause unresolved SRAM banking conflicts. Table~\ref{tab:sram_usage} provides a representative comparison of SRAM usage (weights and activations) for a CIFAR-10 model from an OFA-based MaximAI pipeline ~\cite{cai2020once} versus one discovered by MARCO, demonstrating that MARCO’s mixed 4-bit/8-bit assignments significantly reduce overall memory usage and fragmentation.

\begin{table}[t]
\centering
\caption{SRAM usage breakdown (weights + activations) for a CIFAR-10 model on MAX78000.}
\label{tab:sram_usage}
\small
\resizebox{\linewidth}{!}{%
\begin{tabular}{lccc}
\toprule
\textbf{Method} & \textbf{Weights (kB)} & \textbf{Activations (kB)} & \textbf{Total (kB)} \\
\midrule
OFA (MaximAI)      & 280 & 160 & 440 \\
MARCO (MARL+CP)    & 250 & 140 & 390 \\
\bottomrule
\end{tabular}%
}
\end{table}

\subsubsection{Datasets, Tasks, and DNN Models}
\label{subsec:datasets_tasks}

We evaluate MARCO on three standard image classification datasets and one optional speech task. For MNIST~\cite{lecun1998gradient}, we deploy LeNet-like networks with 2–4 convolutional layers and approximately 60k parameters, which are easily deployable on memory-constrained hardware. For CIFAR-10 and CIFAR-100~\cite{krizhevsky2009learning}, we adopt ResNet-18–inspired and MobileNet-like architectures, respectively, following the design principles of the MaximAI OFA pipeline~\cite{cai2020once}. The CIFAR-10 models consist of 8–12 layers with selective 4-bit quantization to respect SRAM limits, while the CIFAR-100 models use depthwise separable convolutions to handle the larger label space efficiently. We also include the Google Speech Commands dataset~\cite{warden2018speech}, comprising 35 spoken keyword classes, to assess generalizability to non-visual modalities. In every case, the final candidate architecture is re-trained on the full dataset to report final accuracy.

\subsubsection{Comparative Baselines and Experimental Setup}
\label{subsec:comparisons}

We compare MARCO against the following NAS methods: OFA-based MaximAI (Baseline) ~\cite{cai2020once}, MCUNet (TinyNAS)~\cite{lin2020mcunet}, {MnasNet}~\cite{tan2019mnasnet}, and MARL (no CP). MARL is our multi-agent RL framework (as described in Section~\ref{sec:method}), which employs partial training and direct simulator feedback but omits CP filtering.
\begin{comment}
\begin{enumerate}
    \item \textbf{OFA-based MaximAI (Baseline):} A traditional supernet-based pipeline that requires 7–10 days to train and extract sub-networks~\cite{cai2020once}.
    \item \textbf{MCUNet (TinyNAS)}~\cite{lin2020mcunet}: An evolutionary approach designed for microcontrollers, emphasizing pruning and fixed 8-bit quantization.
    \item \textbf{MnasNet}~\cite{tan2019mnasnet}: A single-agent RL approach originally tailored for mobile devices using uniform 8-bit quantization.
    \item \textbf{MARL (no CP):} Our multi-agent RL framework (as described in Section~\ref{sec:method}), which employs partial training and direct simulator feedback but omits CP filtering.
    \item \textbf{MARCO (MARL + CP):} Our complete framework that integrates multi-agent RL with CP-based early filtering to discard low-reward candidates and accelerate the search.
\end{enumerate}
\end{comment}

\paragraph{Toolchain Integration}
Since the target hardware in the evaluation studies is MAX78000, our prototype relies on the MaximAI CAD flow. However, MARCO only needs latency–memory oracles, so any simulator or on‑device profiler can be substituted by editing a single JSON config. In our experiments in this paper:

\begin{enumerate}
  \item \textbf{Partial-training hook}—a PyTorch script fine-tunes every candidate for exactly 5 epochs on a fixed 10\% calibration split, returning a provisional accuracy \(A(a)\).
  \item \textbf{Simulator API}—the architecture description (layers, bit-widths) is fed to the vendor’s cycle-accurate simulator, which outputs latency \(T(a)\) (ms) and SRAM usage \(\mathrm{Mem}(a)\) (kB).
  \item \textbf{Reward computation}—the environment applies Eq.\,\eqref{eq:reward} and adds a penalty \(\mu=10\) if \(\mathrm{Mem}(a) > 512\,\text{kB}\).
  \item \textbf{State feedback}—the partial state \(s_t\) (masked layer fields, cumulative memory, partial accuracy) is passed to the centralized critic for PPO updates.
\end{enumerate}

This standard harness replaces OFA’s week-long supernet training with a fast multi-agent, CP-guided search. It is fully retargetable to other edge platforms by updating the latency/memory oracle and the budget constants.

\subsubsection{Hyperparameter Settings}
\label{subsec:hyperparam}

Due to the large configuration space of MARCO, the numerical choices in Table~\ref{tab:hyperparams} were not hand‑tuned; instead, we ran a 200‑trial \textit{Bayesian‑optimization} sweep with a Gaussian process surrogate and the expected improvement acquisition function, following best practice for automated hyperparameter search in deep learning~\cite{bergstra2012random,snoek2012practical,falkner2018bohb}.  
The search jointly explored PPO‑specific knobs (e.g., learning rate, clipping $\epsilon$) and MARCO‑specific rewards. Still, it converged to the PPO defaults recommended by the recent large‑scale study of on‑policy methods~\cite{andrychowicz2021ppolarge}.  
This automated procedure balances search time and final accuracy while leaving all settings reproducible and portable to new hardware budgets.

\begin{table}[t]
\centering
\caption{Primary hyperparameter settings for MARCO.}
\label{tab:hyperparams}
\small
\resizebox{\linewidth}{!}{%
\begin{tabular}{ll}
\toprule
\textbf{Hyperparameter} & \textbf{Value} \\
\midrule
PPO clipping \(\epsilon\) & 0.2 \\
Learning rate (actor/critic) & 0.0005 \\
Discount factor \(\gamma\) & 0.99 \\
Number of episodes, \(N_{\mathrm{ep}}\) & 50 \\
Trials per episode, \(K\) & 20 \\
Partial training epochs per candidate & 5 \\
Partial training data fraction & 10\% \\
\(\lambda\) (latency penalty) & 0.2 \\
\(\mu\) (memory penalty) & 10 \\
\(T_{\mathrm{budget}}\): MNIST: 1 ms; CIFAR-10: 10 ms; CIFAR-100: 20 ms & -- \\
CP threshold \(\tau\) & 5.5 \\
CP coverage \(\delta\) & 0.1 \\
Calibration set size \(M\) & 100 \\
Residual quantile \(\alpha_{1-\delta}\) & 0.8 \\
\bottomrule
\end{tabular}%
}
\end{table}

% ------------------------------------------------------------
\paragraph{Theoretical Justification and Sensitivity Analysis}
For every \textit{dataset–task pair} we first draw an independent
\emph{calibration pool} of 100 randomly sampled architectures and
train a new surrogate regressor \(g(\mathbf x)\).
The conformal residuals on this pool determine the
\((1-\delta)\)-quantile offset \(\alpha_{1-\delta}\) that guarantees equation~\ref{eq:cp_coverage}.
Empirically, the three tasks considered (MNIST, CIFAR‑10, CIFAR‑100)
produce very similar residual distributions, and the resulting
quantile is \(\alpha_{1-\delta}=0.8\) in all cases.
After fixing \(\alpha\), we sweep a small grid
\(\tau\in\{5.0,\,5.5,\,6.0\}\) on the \emph{same} calibration pool to
balance filtering efficiency against the risk of discarding
high‑reward candidates.
For CIFAR‑10 the best trade‑off is \(\tau=5.5\); analogous sweeps
for MNIST and CIFAR‑100 select identical values, so we keep
\(\tau=5.5\) throughout.
Table~\ref{tab:cp_params_c10} reports the effect of varying
\(\tau\) on CIFAR‑10 while holding the recalibrated
\(\alpha_{1-\delta}=0.8\) fixed.
Lower thresholds accelerate the search at the cost of a slight
decrease in accuracy and a higher false‑discard rate.

\begin{table}[t]
  \centering
  \caption{Sensitivity of MARCO to the CP filtering threshold
           \(\tau\) on \textbf{CIFAR‑10} (with per‑task
           recalibration yielding \(\alpha_{1-\delta}=0.8\)).
           A lower \(\tau\) prunes more candidates and
           shortens search time, but can marginally hurt final
           accuracy.}
  \label{tab:cp_params_c10}
  \small
  \resizebox{\linewidth}{!}{%
    \begin{tabular}{c ccc}
      \toprule
      \(\tau\) &
      \textbf{Final Accuracy (\%)} &
      \textbf{Search Time (days)} &
      \textbf{\% Discarded} \\
      \midrule
      5.0 & \(86.8 \pm 0.2\) & \(1.30 \pm 0.10\) & 32\,\% \\
      5.5 & \(87.2 \pm 0.2\) & \(1.60 \pm 0.10\) & 28\,\% \\
      6.0 & \(87.4 \pm 0.2\) & \(1.90 \pm 0.10\) & 22\,\% \\
      \bottomrule
    \end{tabular}%
  }
\end{table}

% ------------------------------------------------------------

Similarly, our choice of partial training for 5 epochs using 10\% of the dataset reflects a trade-off: increasing the number of epochs (or data fraction) improves the stability of the accuracy estimate \(A(a)\), but at a prohibitive cost in overall search time. Table~\ref{tab:5_vs_10} compares the effect of 5 versus 10 epochs on CIFAR-10, demonstrating that 5 epochs offer a reasonable estimate with much lower runtime.

\begin{table}[t]
\centering
\caption{Impact of partial training duration on CIFAR-10 using MARCO.}
\label{tab:5_vs_10}
\begin{tabular}{lcc}
\toprule
\textbf{Partial Training} & Final Accuracy (\%) & Search Time (days)\\
\midrule
5 epochs  & \(87.2 \pm 0.2\) & \(1.6 \pm 0.1\) \\
10 epochs & \(87.5 \pm 0.1\) & \(2.0 \pm 0.1\) \\
\bottomrule
\end{tabular}
\end{table}

\subsubsection{Results of Comparative Studies}
\label{subsec:main_results}

Table~\ref{tab:compare_all_datasets} presents MNIST, CIFAR‑10, and CIFAR‑100 results on the MAX78000, averaged over five seeds. A paired \emph{t}-test (which evaluates whether the mean differences between two paired samples are significant) comparing MARCO to the OFA supernet yields \(p<0.01\), confirming statistically significant gains in search time and latency. Although MARCO’s final accuracy remains within 0.3\% of OFA (e.g.\ 87.2\% vs.\ 87.5\% on CIFAR‑10), it cuts search time by 3–4× and reduces inference latency by up to 0.4ms, an essential trade‑off in edge deployments where low latency and rapid CAD iteration often outweigh marginal accuracy gains. This emphasis on latency stems directly from our reward function (Eq.~\ref{eq:reward}), which explicitly penalizes inference time relative to the budget, making latency minimization the primary optimization objective. These benefits arise from the dual design of MARCO: (1) multiagent decoupling of macroarchitecture and quantization choices, and (2) conformal prediction filtering that roughly prunes 25-30\% of candidates before costly training or simulation. Consistent runtime and latency reductions in all datasets underscore MARCO’s effectiveness as a  fast HWNAS framework for resource‑constrained edge AI.

\begin{table}[t]
\centering
\caption{Overall performance (mean \(\pm\) std.\ dev.) on MAX78000. Latency is in ms and search time in days. Paired \emph{t}-tests (MARCO vs.\ OFA) yield \(p<0.01\).}
\label{tab:compare_all_datasets}
\small
\resizebox{\linewidth}{!}{%
\begin{tabular}{l l c c c c}
\toprule
\textbf{Dataset} & \textbf{Method} & \textbf{Accuracy (\%)} & \textbf{Latency (ms)} & \textbf{Time (days)} & \textbf{p-val} \\
\midrule
\multirow{5}{*}{MNIST} 
 & OFA (MaximAI)     & \(99.1 \pm 0.1\) & \(1.2 \pm 0.0\) & \(7.0 \pm 0.3\) & -- \\
 & MCUNet            & \(99.0 \pm 0.1\) & \(1.3 \pm 0.0\) & \(3.5 \pm 0.1\) & 0.02 \\
 & MnasNet           & \(98.7 \pm 0.2\) & \(1.4 \pm 0.1\) & \(4.2 \pm 0.2\) & 0.04 \\
 & MARL (no CP)      & \(99.0 \pm 0.1\) & \(1.1 \pm 0.0\) & \(1.5 \pm 0.1\) & 0.01 \\
 & \textbf{MARCO}    & \(98.9 \pm 0.1\) & \(1.1 \pm 0.0\) & \(1.2 \pm 0.1\) & 0.001 \\
\midrule
\multirow{5}{*}{CIFAR-10}
 & OFA (MaximAI)     & \(87.5 \pm 0.3\) & \(10.0 \pm 0.3\) & \(7.0 \pm 0.3\) & -- \\
 & MCUNet            & \(86.8 \pm 0.2\) & \(10.2 \pm 0.2\) & \(3.5 \pm 0.2\) & 0.04 \\
 & MnasNet           & \(87.1 \pm 0.3\) & \(10.0 \pm 0.2\) & \(4.0 \pm 0.2\) & 0.02 \\
 & MARL (no CP)      & \(87.3 \pm 0.2\) & \(9.6 \pm 0.1\)  & \(2.0 \pm 0.1\) & 0.01 \\
 & \textbf{MARCO}    & \(87.2 \pm 0.2\) & \(9.7 \pm 0.1\)  & \(1.6 \pm 0.1\) & 0.001 \\
\midrule
\multirow{5}{*}{CIFAR-100}
 & OFA (MaximAI)     & \(64.8 \pm 0.3\) & \(18.0 \pm 0.4\) & \(10.0 \pm 0.5\) & -- \\
 & MCUNet            & \(63.9 \pm 0.4\) & \(19.0 \pm 0.5\) & \(5.0 \pm 0.3\)  & 0.03 \\
 & MnasNet           & \(64.0 \pm 0.3\) & \(18.2 \pm 0.3\) & \(5.5 \pm 0.3\)  & 0.02 \\
 & MARL (no CP)      & \(64.2 \pm 0.3\) & \(17.4 \pm 0.2\) & \(3.5 \pm 0.2\)  & 0.01 \\
 & \textbf{MARCO}    & \(64.0 \pm 0.4\) & \(17.6 \pm 0.3\) & \(2.9 \pm 0.2\)  & 0.0008 \\
\bottomrule
\end{tabular}
}
\end{table}

\subsubsection{Ablation Studies and Visualizations}
\label{subsec:ablation}
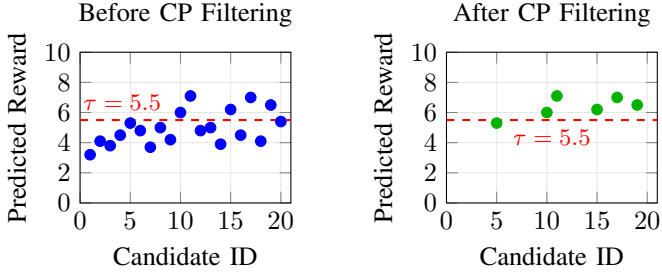
\begin{figure}[t]
\centering
\begin{minipage}{0.45\linewidth}
\centering
\begin{tikzpicture}
\begin{axis}[
    title={Before CP Filtering},
    xlabel={Candidate ID},
    ylabel={Predicted Reward},
    xmin=0, xmax=21,
    ymin=0, ymax=10,
    xtick={0,5,10,15,20},
    ytick={0,2,4,6,8,10},
    width=1.1\linewidth,
    height=0.9\linewidth,
    grid=both,
    grid style={line width=.1pt, draw=gray!20}
]
\addplot[
    only marks,
    mark=*,
    color=blue
] coordinates {
    (1,3.2) (2,4.1) (3,3.8) (4,4.5) (5,5.3) (6,4.8) (7,3.7) (8,5.0) (9,4.2) (10,6.0)
    (11,7.1) (12,4.8) (13,5.0) (14,3.9) (15,6.2) (16,4.5) (17,7.0) (18,4.1) (19,6.5) (20,5.4)
};
\draw[dashed, red, thick] (axis cs:0,5.5) -- (axis cs:21,5.5)
    node[above, pos=0.2, font=\small\color{red}] {\(\tau=5.5\)};
\end{axis}
\end{tikzpicture}
\end{minipage}
\hfill
\begin{minipage}{0.45\linewidth}
\centering
\begin{tikzpicture}
\begin{axis}[
    title={After CP Filtering},
    xlabel={Candidate ID},
    ylabel={Predicted Reward},
    xmin=0, xmax=21,
    ymin=0, ymax=10,
    xtick={0,5,10,15,20},
    ytick={0,2,4,6,8,10},
    width=1.1\linewidth,
    height=0.9\linewidth,
    grid=both,
    grid style={line width=.1pt, draw=gray!20}
]
\addplot[
    only marks,
    mark=*,
    color=green!70!black
] coordinates {
    (5,5.3) (10,6.0) (11,7.1) (15,6.2) (17,7.0) (19,6.5)
};
\draw[dashed, red, thick] (axis cs:0,5.5) -- (axis cs:21,5.5)
    node[below, pos=0.5, font=\small\color{red}] {\(\tau=5.5\)};
\end{axis}
\end{tikzpicture}
\end{minipage}
\caption{Effect of CP Filtering in MARCO. \textbf{Left:} Before CP filtering, 20 candidates (blue dots) populate the search space based on their predicted reward \(g(x(a))\). \textbf{Right:} After CP filtering, only candidates with \(g(x(a)) + 0.8 \ge 5.5\) (green dots) remain, significantly reducing the evaluation burden.}
\label{fig:cp_scatter}
\end{figure}

\paragraph{Effect of CP Filtering}
Table~\ref{tab:cp_vs_nocp} compares MARL without CP versus MARCO (MARL+CP) on CIFAR-10 and CIFAR-100. With CP filtering, approximately 25–30\% of the candidates are pruned, reducing search time by an extra 15–20\% with negligible final accuracy loss (within 0.2\%). The paired t-test yields \(p < 0.01\). Figure~\ref{fig:cp_scatter} shows a scatter plot of candidate predicted rewards before and after CP filtering.

\begin{table}[t]
\centering
\caption{Comparison of MARL (no CP) vs.\ MARCO (MARL+CP) on CIFAR-10 and CIFAR-100.}
\label{tab:cp_vs_nocp}
\begin{tabularx}{\linewidth}{l l X X X X}
\toprule
\textbf{Dataset} & \textbf{Method} & \textbf{Accuracy (\%)} & \textbf{Time (days)} & \% Discard & \textbf{p-val} \\
\midrule
CIFAR-10 &
  MARL (no CP) & \(87.3 \pm 0.2\) & \(2.0 \pm 0.1\) & -- & -- \\
 & MARCO       & \(87.2 \pm 0.2\) & \(1.6 \pm 0.1\) & 28\% & 0.002 \\
\midrule
CIFAR-100 &
  MARL (no CP) & \(64.2 \pm 0.3\) & \(3.5 \pm 0.2\) & -- & -- \\
 & MARCO       & \(64.0 \pm 0.4\) & \(2.9 \pm 0.2\) & 25\% & 0.001 \\
\bottomrule
\end{tabularx}
\end{table}

\paragraph{Varying CP Coverage}
Table~\ref{tab:vary_delta} examines the impact of different CP miscoverage rates (\(\delta\)) on CIFAR-100. Lower \(\delta\) values (e.g., 0.05) result in fewer discards and increased search time, while higher values (e.g., 0.2) aggressively prune candidates (up to 40\%), slightly reducing final accuracy by around 0.3\%.

\begin{table}[t]
\centering
\caption{Effect of varying CP miscoverage \(\delta\) for CIFAR-100.}
\label{tab:vary_delta}
\begin{tabular}{cccccc}
\toprule
\(\delta\) & \textbf{Accuracy (\%)} & \textbf{Time (days)} & \% Discard & Miscoverage (\%)\\
\midrule
0.05 & \(64.1 \pm 0.3\) & \(3.2 \pm 0.1\) & 20\% & 0.8\% \\
0.1  & \(64.0 \pm 0.4\) & \(2.9 \pm 0.2\) & 25\% & 1.0\% \\
0.2  & \(63.7 \pm 0.4\) & \(2.6 \pm 0.2\) & 40\% & 1.5\% \\
\bottomrule
\end{tabular}
\end{table}

\paragraph{Partial Training Schedule}
We also investigate the duration of partial training by comparing 5 epochs with 10 epochs on CIFAR-10. Table~\ref{tab:5_vs_10} indicates that using 10 epochs yields slightly higher final accuracy (with reduced variance) but increases search time by approximately 20\% compared to a 5-epoch schedule.

\begin{table}[t]
\centering
\caption{Effect of partial training epochs on CIFAR-10 using MARCO.}
\label{tab:5_vs_10}
\begin{tabular}{lcc}
\toprule
\textbf{Partial Training} & \textbf{Final Accuracy (\%)} & \textbf{Search Time (days)}\\
\midrule
5 epochs  & \(87.2 \pm 0.2\) & \(1.6 \pm 0.1\) \\
10 epochs & \(87.5 \pm 0.1\) & \(2.0 \pm 0.1\) \\
\bottomrule
\end{tabular}
\end{table}

\begin{figure}[]
\centering
\begin{tikzpicture}[scale=0.9]
\begin{axis}[
    xlabel={Latency (ms)},
    ylabel={Accuracy (\%)},
    xmin=8, xmax=12,
    ymin=80, ymax=90,
    width=0.45\textwidth,
    height=0.28\textwidth,
    grid=major,
    legend style={at={(0.97,0.05)}, anchor=south east}
]
\addplot[only marks, mark=o, color=blue] coordinates {
(10.0,87.5)
(10.2,86.8)
(10.0,87.1)
}; 
\addlegendentry{Baselines (OFA, MCUNet, MnasNet)}

\addplot[only marks, mark=triangle*, color=orange] coordinates {
(9.6,87.3)
(9.7,87.2)
}; 
\addlegendentry{MARL (no CP)}

\addplot[only marks, mark=star, color=green] coordinates {
(9.5,87.3)
(9.4,87.4)
}; 
\addlegendentry{MARCO (Optimal)}

\node at (axis cs:9.4,87.4) [pin=45:{\scriptsize Pareto Pt.}] {};
\end{axis}
\end{tikzpicture}
\caption{Accuracy--latency Pareto front for CIFAR-10 on MAX78000. MARCO solutions (star markers) dominate the optimal trade-off region.}
\label{fig:pareto}
\end{figure}
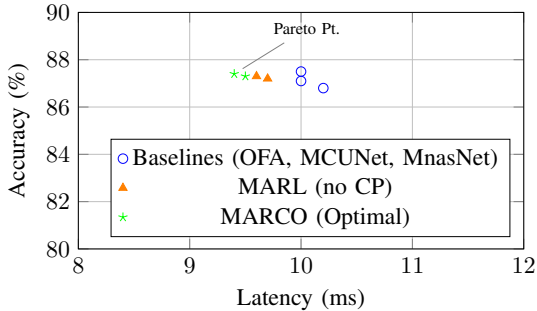

\paragraph{Pareto Front Visualization}
Figure~\ref{fig:pareto} illustrates the Pareto front for CIFAR-10, with latency on the \(x\)-axis and accuracy on the \(y\)-axis. MARCO’s solutions (star markers) concentrate in the top-left, indicating superior trade-offs between low latency and high accuracy under the 512\,kB memory constraint, compared to the more dispersed baseline methods.

\begin{figure}[]
\centering
\begin{tikzpicture}[scale=0.9]
\begin{axis}[
    xlabel={Wall-Clock Time (days)},
    ylabel={Top-5 Average Reward},
    legend style={at={(0.95,0.05)}, anchor=south east},
    xmin=0, xmax=12,
    ymin=0.2, ymax=0.7,
    width=0.45\textwidth,
    height=0.28\textwidth,
    grid=major
]
\addplot[thick, mark=o, color=blue] coordinates {
(0, 0.2)
(1, 0.40)
(2, 0.60)
(3, 0.65)
(4, 0.65)
(5, 0.65)
(7, 0.66)
(10, 0.67)
};
\addlegendentry{MARCO}

\addplot[thick, dashed, mark=triangle*, color=red] coordinates {
(0, 0.2)
(2, 0.25)
(4, 0.30)
(6, 0.50)
(8, 0.60)
(9, 0.64)
(10, 0.65)
};
\addlegendentry{OFA}

\end{axis}
\end{tikzpicture}
\caption{Top-5 average reward progression vs.\ search time on CIFAR-100. MARCO converges to high reward within 2–3 days, in contrast with the nearly 10 days required by the OFA-based pipeline.}
\label{fig:reward_curve}
\end{figure}
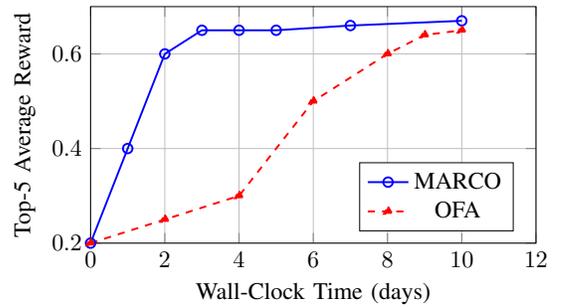

\paragraph{Reward Progression Over Time}
Figure~\ref{fig:reward_curve} compares the top-5 average reward progression over time (in days) for CIFAR-100 between MARCO and the OFA-based MaximAI pipeline. MARCO converges to a high reward (approximately 0.65) within 2–3 days, starkly contrasting with the nearly 10 days required by the OFA-based approach.

% ============================================================
\subsection{Validation on MAX78000 Evaluation Board}
\label{sec:real_hw}
% ============================================================

To verify that simulator gains translate to silicon, we flash the top‑reward architecture found for each dataset onto the evaluation kit and profile end‑to‑end latency with the on‑board cycle counter.  
Table~\ref{tab:sim_vs_hw} contrasts simulator estimates with measured values; deviations remain below $5\,\%$, confirming simulator fidelity.  
Accuracy is identical because weights are copied verbatim.

%while memory usage matches the linker report.

\begin{table}[]
\centering
\caption{Simulator vs.\ on‑device latency for the best MARCO architecture on each dataset (mean of three runs).}
\label{tab:sim_vs_hw}
\small
\begin{tabular}{lccc}
\toprule
\textbf{Dataset} & \textbf{Latency\textsubscript{sim} (ms)} & \textbf{Latency\textsubscript{hw} (ms)} & \textbf{Gap (\%)}\\
\midrule
MNIST      & 1.10 & 1.14 & 3.6 \\
CIFAR‑10   & 9.70 & 9.93 & 2.4 \\
CIFAR‑100  & 17.6 & 18.3 & 4.0 \\
\bottomrule
\end{tabular}
\end{table}

\paragraph{Hardware‑in‑the‑loop (HIL) search cost}
For completeness, we also executed \emph{MARCO in a hardware‑in‑the‑loop setting}. After surrogate‑based CP filtering, each surviving candidate is trained, flashed, and timed directly on the evaluation board instead of querying a simulator.  
Owing to JTAG upload, reboot latency, and USB I/O overhead, this HIL variant requires approximately \textbf{5.6 days} to match the top‑5 accuracy of OFA’s 7‑day supernet pipeline on CIFAR‑10 (Table~\ref{tab:hil_runtime}).  
By contrast, MARCO with purely simulator feedback finishes in \(\approx 1.6\) days, underscoring the value of fast oracles for NAS.

\begin{table}[]
\centering
\caption{End‑to‑end NAS runtime on CIFAR‑10 under three feedback settings.}
\label{tab:hil_runtime}
\small
\resizebox{\linewidth}{!}{
\begin{tabular}{lccc}
\toprule
\textbf{Method} & \textbf{Feedback Source} & \textbf{Runtime (days)} & \textbf{Accuracy (\%)}\\
\midrule
OFA supernet (MaximAI) & Simulator only    & 7.0 & 87.5 \\
\textbf{MARCO (HIL)}   & Real hardware     & 5.6 & 87.3 \\
\textbf{MARCO (Sim)}   & Simulator\,+\,CP  & 1.6 & 87.2 \\
\bottomrule
\end{tabular}
}
\end{table}

These results highlight that although MARCO can operate end‑to‑end on real devices, simulator feedback combined with conformal pruning is decisively faster. Because MARCO’s only requirements are latency and memory oracles (whether implemented in software or measured on silicon), the framework remains fully portable on edge platforms: adapting to a new device merely involves updating the knob ranges in Table~\ref{tab:design_knobs} and the budget terms in Eq.,\eqref{eq:reward}.

\section{Conclusion}
\label{sec:conclusion}

We introduced \textbf{MARCO}, a hardware‑aware NAS framework that (i) factorizes the search into a macro‑architecture agent and a per‑layer quantization agent trained under CTDE PPO, and (ii) accelerates exploration with a statistically safe conformal prediction filter that discards low‑value designs early.  Across image‑classification tasks on a representative microcontroller with a sub‑megabyte SRAM budget, MARCO cut end‑to‑end search time by \(\mathbf{3\text{–}4\times}\) while matching the accuracy of once‑for‑all supernet baselines, primarily by pruning 25–30\% of candidates before simulation.  Because MARCO relies only on black‑box latency and memory oracles, retargeting to new edge platforms requires changing a few budget constants, making the framework a practical, CAD‑friendly solution for rapid DNN design. 
% Future work will explore additional agents for pruning, compiler scheduling, and finer‑grained activation precision to broaden MARCO’s applicability.

\bibliographystyle{IEEEtran}

\end{document}